# A Behavioral Distance for Fuzzy-Transition Systems

Yongzhi Cao, *Member, IEEE*, Huaiqing Wang, Sherry X. Sun, and Guoqing Chen

*Abstract*—In contrast to the existing approaches to bisimulation for fuzzy systems, we introduce a behavioral distance to measure the behavioral similarity of states in a nondeterministic fuzzy-transition system. This behavioral distance is defined as the greatest fixed point of a suitable monotonic function and provides a quantitative analogue of bisimilarity. The behavioral distance has the important property that two states are at zero distance if and only if they are bisimilar. Moreover, for any given threshold, we find that states with behavioral distances bounded by the threshold are equivalent. In addition, we show that two system combinators—parallel composition and product— are non-expansive with respect to our behavioral distance, which makes compositional verification possible.

## I. INTRODUCTION

ONE of the most important contributions of concurrency theory to computer science is the concept of bisimulation (see, for example, [1]–[3] and the bibliographies therein). It expresses when two systems can behave in the same way in the sense that one system simulates the other and vice-versa; intuitively, two systems are bisimilar if they match each other's moves. In addition to testing behavioral equivalence, bisimulation allows one to reduce the state space of a system by combining bisimilar states.

Recently, bisimulation techniques have been introduced to fuzzy automata, or more generally, fuzzy systems. Two general approaches can be recognized in the existing literature. One is based on a binary relation on the state space of a fuzzy system such that related states have exactly the same possibility degree of making a transition into every class of related states [4]–[8]. Roughly speaking, this approach is a fuzzy analogue of the classical probabilistic bisimulation initiated by Larsen and Skou [9]. More concretely, in [4], Petković introduced the notion of congruence for fuzzy automata by following the algebraic theory of classical automata. In fact, such a congruence is nothing other than a bisimulation. Based on the concept of congruence, an improved minimization algorithm for fuzzy automata has been developed. In [5], Buchholz has put forward a general definition of bisimulation for weighted automata over a generic semiring. By instantiating the semiring to the closed unit interval $[0, 1]$ with binary operations $\max$ and $\min$, Sun *et al.* addressed the forward and backward bisimulations for fuzzy automata [6]. More recently, the Cao *et al.* investigated bisimulation for deterministic and nondeterministic fuzzy systems, which may be infinite-state or infinite-event, in [7] and [8], respectively, and applied it to the specification of nondeterministic fuzzy discrete-event systems [8].

The other approach, from Ćirić and his colleagues [10]–[15], is based on a fuzzy relation on the state space. They introduced two types of simulations (forward and backward) and four types of bisimulations (forward, backward, forward-backward, and backward-forward) for fuzzy automata, which are all defined as fuzzy relations [13], [15]. In particular, they proved that the greatest (forward) bisimulation on a fuzzy automaton is a fuzzy equivalence relation. Furthermore, the state reduction of fuzzy automata and some related algorithms for computing the greatest simulations and bisimulations have been well developed. Remarkably, Ćirić *et al.* have found that the state reduction problem for fuzzy automata is closely related to the problem of solving certain systems of fuzzy relation equations [10], [12], [14], which provides a new insight into the theory of fuzzy relational equations and inequalities.

For the first approach, we observe that the bisimulation based on a crisp relation, in which states are either bisimilar or not, is not a robust concept, since states that used to be bisimilar may not be anymore or vice versa if some of the possibility degrees change slightly. This is particularly unfortunate for the reason that such possibility degrees are often obtained experimentally, or are given as approximations. Therefore, it may not make sense to say that two states are exactly bisimilar. In terms of robustness, the bisimulation based on a fuzzy relation is much better than the exact bisimulation. This advantage leads to many better results in the state reduction of fuzzy automata [12], [14].

Inspired by earlier work on probabilistic concurrent systems [16]–[21], in the present paper we exploit a pseudo-ultrametric to measure the similarity of states in a (nondeterministic) fuzzy-transition system (FTS). A pseudo-ultrametric is a function that yields a nonnegative real number distance for each pair of states. Such a pseudo-ultrametric gives rise to a quantitative analogue of exact bisimilarity in that the behavioral distance between states captures the similarity of the behavior of those states. The smaller the behavioral distance, the more the states behave similarly. In particular, the behavioral distance between states is $0$ if and only if they are exactly bisimilar. Moreover, for any given threshold, we can partition the state space such that the behavioral distance between the states in the same block is bounded by the threshold. In other words, states with behavioral distances bounded are

This work was supported by NSFC under Grants 60973004, 61170299, and 70890080, by MOST under Grants 2009CB320701 and 2010CB328103, and by SRG from City University of Hong Kong under Grant 7002630.

Y. Cao is with the Institute of Software, School of Electronics Engineering and Computer Science, Peking University, Beijing 100871, China and the Key Laboratory of High Confidence Software Technologies, Ministry of Education (e-mail: caoyz@pku.edu.cn).

H. Wang and S. X. Sun are with the Department of Information Systems, City University of Hong Kong, Kowloon, Hong Kong (e-mails: iswang@cityu.edu.hk; sherry.sun@cityu.edu.hk).

G. Chen is with the School of Economics and Management, Tsinghua University, Beijing 100084, China (e-mail: chengq@sem.tsinghua.edu.cn).

equivalent. Technically, we define the behavioral distance as the greatest fixed point of a suitable monotonic function, while the post-fixed points of the monotonic function provide a characterization of exact bisimulation. We also show that two system combinators—parallel composition and product—are non-expansive with respect to our behavioral distance. This allows us to use the behavioral distance for compositional reasoning.

Some related research should be distinguished before introducing the organization of the paper. Broadly speaking, our behavioral distance is a fuzzy analogue of the pseudo-metric used in some probabilistic systems [16]–[21]. Nevertheless, building such an analogue is not trivial, and we have to develop a whole new framework. For example, our lifting of pseudo-ultrametrics from states to possibility distributions answers an open problem on the existence of the fuzzy analogue of Kantorovich metric raised by Repovš *et al.* in [22]. Our approach is rather different from the one by Ćirić *et al.* [10]–[15] in at least two aspects: One is that we are using the pseudo-ultrametric, which is related in spirit to the Hutchinson metric and the Hausdorff distance, to provide a robust and quantitative notion of behavioral equivalence, while the second approach is dependent on finding a solution to a particular system of fuzzy relation equations or inequalities. The main feature of the latter is the intensive use of fuzzy relation calculus and systems of fuzzy relation equations and inequalities. The other difference is that the underlying systems addressed by the second approach are deterministic in the sense that the next possibility distribution is determined by the current state and event, while we are concerned with more general nondeterministic fuzzy systems. It should be noted that nondeterminism is essential for modeling scheduling freedom, implementation freedom, the external environment, and incomplete information (see, for example, [23]).

The rest of the paper is organized as follows. In Section II, we collect a few necessary notations and notions of fuzzy sets and FTSs. Section III embarks upon the development of behavioral distance. It starts by lifting a pseudo-ultrametric from states to possibility distributions. Based on the lifting, we then define a function on a set of pseudo-ultrametrics and discuss the monotony of the function. Thanks to Tarski's fixed point theorem, we get the greatest fixed point of the function and define it as our behavioral distance. In Section IV, after establishing a lifting of relation from states to possibility distributions, we justify the soundness of the behavioral distance by disclosing the relationship between the distance and bisimilarity. In the subsequent section, we investigate the non-expansiveness of the behavioral distance with respect to the parallel composition and product operators. The paper is concluded in Section VI with a brief discussion of future work.

## II. FUZZY-TRANSITION SYSTEMS

In this section, we recall some basic notions of fuzzy sets and FTSs.

Let $X$ be a universal set. A *fuzzy subset* of $X$ (or simply *fuzzy set* [24]), $\mu$, is defined by a function assigning to each element $x$ of $X$ a value $\mu(x)$ in the closed unit interval $[0, 1]$. Such a function is called a *membership function*; the value $\mu(x)$ characterizes the degree of membership of $x$ in $\mu$. A fuzzy subset of $X$ can be used to formally represent a possibility distribution on $X$.

The *support* of a fuzzy set $\mu$ is a crisp set defined as $\mathrm{supp}(\mu) = \{x \in X : \mu(x) > 0\}$. Whenever $\mathrm{supp}(\mu)$ is finite, say $\mathrm{supp}(\mu) = \{x_1, x_2, \ldots, x_n\}$, we may write the fuzzy set $\mu$ in Zadeh's notation as follows:

$$\mu = \frac{\mu(x_1)}{x_1} + \frac{\mu(x_2)}{x_2} + \cdots + \frac{\mu(x_n)}{x_n}.$$

With this notation, $\frac{1}{x}$, denoted by $\hat{x}$, is a singleton in $X$, i.e., the fuzzy subset of $X$ with membership 1 at $x$ and with zero membership for all the other elements of $X$.

We denote by $\mathcal{F}(X)$ the set of all fuzzy subsets of $X$ (i.e., possibility distributions on $X$) and by $\mathcal{P}(X)$ the power set of $X$. For any $\mu, \eta \in \mathcal{F}(X)$, we say that $\mu$ is contained in $\eta$ (or $\eta$ contains $\mu$), denoted by $\mu \subseteq \eta$, if $\mu(x) \leq \eta(x)$ for all $x \in X$. Notice that $\mu = \eta$ if both $\mu \subseteq \eta$ and $\eta \subseteq \mu$. A fuzzy set is said to be *empty* if its membership function is identically zero on $X$. We use $\emptyset$ to denote the empty fuzzy set.

For any family $\lambda_i$, $i \in I$, of elements of $[0, 1]$, we write $\vee_{i \in I} \lambda_i$ or $\vee \{\lambda_i : i \in I\}$ for the supremum of $\{\lambda_i : i \in I\}$, and $\wedge_{i \in I} \lambda_i$ or $\wedge \{\lambda_i : i \in I\}$ for the infimum. In particular, if $I$ is finite, then $\vee_{i \in I} \lambda_i$ and $\wedge_{i \in I} \lambda_i$ are the greatest element and the least element of $\{\lambda_i : i \in I\}$, respectively.

For any given $c \in [0, 1]$ and $\mu \in \mathcal{F}(X)$, the *scale product* $c \cdot \mu$ of $c$ and $\mu$ is defined by

$$(c \cdot \mu)(x) = c \wedge \mu(x),$$

for each $x \in X$; this is again a fuzzy set. Given $\mu, \eta \in \mathcal{F}(X)$, the *union* of $\mu$ and $\eta$, denoted $\mu \cup \eta$, is defined by the membership function

$$(\mu \cup \eta)(x) = \mu(x) \vee \eta(x)$$

for all $x \in X$. For any $\mu \in \mathcal{F}(X)$ and $U \subseteq X$, the notation $\mu(U)$ stands for $\vee_{x \in U} \mu(x)$.

We now review the concept of FTSs. In [7], an FTS is defined as a four-tuple $(S, A, \delta, s_0)$, where $S$ is the set of states, $A$ is the set of labels, $\delta$ is a mapping from $S \times A$ to $\mathcal{F}(S)$, and $s_0$ is the initial state. Labels in an FTS can represent different things. Typical uses of labels include representing input expected, conditions that must be true to trigger the transition, or actions performed during the transition. If the label set is a singleton, the system is essentially unlabeled, and a simpler definition that omits the labels is possible. Intuitively, if the FTS is in state $s \in S$ and the label $a \in A$ occurs, then it may go into the state $s' \in S$ with possibility degree $\delta(s, a)(s')$. Such an FTS is deterministic in the sense that for each state $s$ and label $a$, only a possibility distribution $\delta(s, a)$ is returned by $\delta$.

Following [8], [25], in this work we address a more general FTS by taking into account nondeterminism. As a result, for each state $s$ and label $a$, more than one possibility distribution may be returned by $\delta$.

*Definition 1:* A *(nondeterministic) fuzzy-transition system* (FTS) is a three-tuple $(S, A, \delta)$, where

(1) $S$ is the set of states,

(2) $A$ is the set of labels, and
(3) $\delta$, called a fuzzy transition function, is a mapping from $S \times A$ to $\mathcal{P}(\mathcal{F}(S))$.

If $(S, A, \delta)$ is an FTS such that $s \in S$, $a \in A$, $\mu \in \mathcal{F}(S)$, and $\mu \in \delta(s, a)$, we write $s \xrightarrow{a} \mu$ and call it a *fuzzy transition*. An FTS is said to be *finite* if both $S$ and $A$ are finite, and *infinite* otherwise. For simplicity, we have excluded the initial state from consideration and will work exclusively with finite FTSs in the paper.

## III. Behavioral Distance

This section, consisting of three subsections, is devoted to quantifying the behavioral similarity between states of an FTS. This quantity, which is based upon the fuzzy transitions derived from the states, meets some desirable metric properties.

### A. Lifting of Pseudo-Ultrametric

In this subsection, we lift a pseudo-ultrametric from states to possibility distributions. Let us first collect some basic notions on pseudo-ultrametric space.

*Definition 2:* Let $X$ be a nonempty universe. A function $d : X \times X \longrightarrow [0, 1]$ is called a *pseudo-ultrametric* on $X$ if for all $x, y, z \in X$,
 (P1) $d(x, x) = 0$,
 (P2) $d(x, y) = d(y, x)$, and
 (P3) $d(x, z) \leq d(x, y) \vee d(y, z)$.
The couple $(X, d)$ is called a *pseudo-ultrametric space*. If the triangle inequality (P3) is weakened by
 (P3') $d(x, z) \leq d(x, y) + d(y, z)$,
then $d$ is called a *pseudo-metric* and $(X, d)$ is a *pseudo-metric space*. At the same time, if (P1) is strengthened by
 (P1') $d(x, y) = 0$ if and only if $x = y$,
that is, $d$ satisfies (P1'), (P2), and (P3'), then $d$ is called a *metric* and $(X, d)$ is a *metric space*.

To simplify notation, we sometimes write $X$ instead of $(X, d)$. Trivially, the constant function that maps any pair $(x, y)$ to 0 is a pseudo-ultrametric. In addition, the discrete metric, where $d(x, y) = 0$ if $x = y$ and $d(x, y) = 1$ otherwise, is a pseudo-ultrametric.

The notion of pseudo-ultrametric has an intuitive interpretation [26]: If $G$ is an edge-weighted undirected graph, all edge weights are in $[0, 1]$, and $d(x, y)$ is the weight of the minimax path between vertices $x$ and $y$ (that is, the maximum weight of an edge, on a path chosen to minimize this maximum weight), then the vertices of the graph, with distance measured by $d$, form a pseudo-ultrametric space, and all finite pseudo-ultrametric spaces can be represented in this way.

A special metric satisfying (P3) appeared in several areas of mathematics in early 20th century (K. Hensel, 1904; R. Baire, 1909; F. Hausdorff, 1914). The corresponding metric space is called *ultrametric* or *non-Archimedean*. A brief review on ultrametrics, including a wide list of the references, can be found in Lemin's paper [27]. Recently, some significant examples of pseudo-metric spaces have arisen in probabilistic transition systems [20], [21], Markov chains [16], [17], and Markov decision processes [18], [19].

For simplicity, we restrict ourself to $[0, 1]$-valued pseudo-ultrametrics. Notice that every pseudo-ultrametric $d'$ with codomain $[0, +\infty)$ can be homeomorphically transformed into a $[0, 1]$-valued pseudo-ultrametric $d$ by defining

$$d(x, y) = \frac{d'(x, y)}{1 + d'(x, y)},$$

for all $x$ and $y$ in $X$ [28].

From now on, we fix a finite FTS $(S, A, \delta)$. Let $\mathcal{D}$ be the set of all pseudo-ultrametrics on $S$. Clearly, $\mathcal{D} \neq \emptyset$. Given a $d \in \mathcal{D}$, we need to lift it to a pseudo-ultrametric on $\mathcal{F}(S)$. The lifting is based on the following observation.

*Lemma 1:* For any $\mu, \eta \in \mathcal{F}(S)$, consider the following system:

$$\begin{cases} \vee_{t \in S} x_{st} = \mu(s), & \forall s \in S \\ \vee_{s \in S} x_{st} = \eta(t), & \forall t \in S \\ x_{st} \geq 0, & \forall s, t \in S \end{cases} \quad (1)$$

Then it has a solution if and only if $\mu(S) = \eta(S)$.
 *Proof:* See Appendix A. ∎

To understand the above lemma and its proof, let us see a simple example.

*Example 1:* Letting $S = \{s, t\}$, $\mu = \frac{0.9}{s} + \frac{0.3}{t}$, and $\eta = \frac{1}{s} + \frac{0.5}{t}$, the system (1) reduces to

$$\begin{cases} x_{ss} \vee x_{st} = 0.9 \\ x_{ts} \vee x_{tt} = 0.3 \\ x_{ss} \vee x_{ts} = 1 \\ x_{st} \vee x_{tt} = 0.5 \\ x_{st} \geq 0, \forall s, t \end{cases}$$

The third equation implies that either $x_{ss} = 1$ or $x_{ts} = 1$. If $x_{ss} = 1$, it contradicts with the first equation; while if $x_{ts} = 1$, it contradicts with the second equation. Therefore, the above system has no solution. However, if we consider $\theta = \frac{0.9}{s} + \frac{0.5}{t}$ and take $x_{st}$ by following the proof of the lemma, it gives rise to a solution to the system (1): $x_{ss} = 0.9, x_{st} = 0.5, x_{ts} = 0.3$, and $x_{tt} = 0$.

With the aid of the above lemma, we can now state the concept of lifting.

*Definition 3:* Let $d \in \mathcal{D}$. For any $\mu, \eta \in \mathcal{F}(S)$, if $\mu(S) \neq \eta(S)$, we define $\hat{d}(\mu, \eta) = 1$; otherwise, we define $\hat{d}(\mu, \eta)$ as the value of the following mathematical programming problem:

$$\begin{aligned}
\text{minimize} \quad & \vee_{s,t \in S} (d(s,t) \wedge x_{st}) \\
\text{subject to} \quad & \vee_{t \in S} x_{st} = \mu(s), \quad \forall s \in S \quad \text{(MP)} \\
& \vee_{s \in S} x_{st} = \eta(t), \quad \forall t \in S \\
& x_{st} \geq 0, \quad \forall s, t \in S
\end{aligned}$$

Let us revisit Example 1. If $d$ is the discrete metric on $S = \{s, t\}$, then it follows readily from Definition 3 that $\hat{d}(\mu, \mu) = \hat{d}(\eta, \eta) = \hat{d}(\theta, \theta) = 0$, $\hat{d}(\mu, \eta) = \hat{d}(\eta, \mu) = 1$, $\hat{d}(\eta, \theta) = \hat{d}(\theta, \eta) = 1$, and $\hat{d}(\mu, \theta) = \hat{d}(\theta, \mu) = 0.5$. In terms of these values, $\hat{d}$ satisfies all the requirements of pseudo-ultrametric. Before exploring the universality of this property, we pause to give two remarks.

*Remark 1:* Whenever $\mu(S) = \eta(S)$, it follows from Lemma 1 that the system (1) has a solution. It implies that the mathematical programming problem (MP) in Definition 3 has a feasible solution. In this case, it is not difficult to observe that there exists an optimal solution to (MP) by considering all the alternatives of $x_{st}$ from the finite set $\{\mu(s) : s \in S\} \cup \{\eta(t) : t \in S\} \cup \{0\}$. Hence, Definition 3 is well-defined.

*Remark 2:* In fact, Definition 3 is inspired by the duality of Kantorovich metric on probability measures. Roughly speaking, the Kantorovich metric and its duality provide a way of measuring the distance between two probability distributions. We refer the reader to [29] for the history of the Kantorovich metric and to [30] for its applications in probabilistic concurrency, image retrieval, data mining, and bioinformatics. Definition 3 is derived from the duality of Kantorovich metric by replacing the sum and product operations therein by max and min, respectively. Most recently, Repovš *et al.* [22] posed an open question: Is there a fuzzy analogue of the Kantorovich metric on the set of probability measures? Our Definition 3 offers up a solution to this problem.

The desirability of Definition 3 is justified by the following fact.

*Theorem 1:* For each $d \in \mathcal{D}$, $\hat{d}$ is a pseudo-ultrametric on $\mathcal{F}(S)$.

*Proof:* See Appendix A. ∎

### B. A Monotonic Function on a Complete Lattice

For later need, we start by discussing the lattice structure on $\mathcal{D}$ and then define a suitable monotonic function on $\mathcal{D}$. To this end, we need to endow $\mathcal{D}$ an order.

*Definition 4:* The order $\preceq$ on $\mathcal{D}$ is defined by

$$d_1 \preceq d_2 \text{ if } d_1(s,t) \geq d_2(s,t) \text{ for all } s,t \in S.$$

It is easy to check that $\preceq$ is indeed a partial order on $\mathcal{D}$.

Recall that a partially ordered set $(X, \leq)$ is called a *complete lattice* if every subset of $X$ has a supremum and an infimum in $(X, \leq)$. We now show that $\mathcal{D}$ endowed with the order specified in Definition 4 forms a complete lattice.

*Lemma 2:* $(\mathcal{D}, \preceq)$ is a complete lattice.

*Proof:* See Appendix A. ∎

For any $d \in \mathcal{D}$, we can get by Theorem 1 a pseudo-ultrametric $\hat{d}$ on possibility distributions. Further, let us extend $\hat{d}$ to a distance measure on the sets of possibility distributions.

Recall that the well-known Hausdorff distance measures how far two subsets of a metric space are from each other. Informally, the Hausdorff distance is the longest distance of either set to the nearest point in the other set. For our purpose, we consider the Hausdorff distance for a pseudo-ultrametric space.

*Definition 5:* Let $(X, d)$ be a pseudo-ultrametric space. For any $x \in X$ and $A \subseteq X$, define

$$d(x, A) = \begin{cases} \wedge_{a \in A} d(x, a), & \text{if } A \neq \emptyset \\ 1, & \text{otherwise.} \end{cases}$$

Further, given a pair $A, B \subseteq X$, the *Hausdorff distance* induced by $d$ is defined as

$$H_d(A,B) = \begin{cases} 0, & \text{if } A = B = \emptyset \\ [\vee_{a \in A} d(a, B)] \vee [\vee_{b \in B} d(b, A)], & \text{otherwise.} \end{cases}$$

As expected, $H_d$ has the following property.

*Lemma 3:* If $d$ is a pseudo-ultrametric on $X$, then $H_d$ is a pseudo-ultrametric on $\mathcal{P}(X)$.

*Proof:* It follows directly from Definition 5, and we thus omit the proof. ∎

Observe that by Theorem 1 any $d \in \mathcal{D}$ induces a pseudo-ultrametric $\hat{d}$ on $\mathcal{F}(S)$; further, it yields a pseudo-ultrametric $H_{\hat{d}}$ on $\mathcal{P}(\mathcal{F}(S))$ by the Hausdorff distance. Based on $H_{\hat{d}}$, let us define a function $\Delta$ on $\mathcal{D}$.

*Definition 6:* The function $\Delta : \mathcal{D} \longrightarrow \mathcal{D}$ is defined as follows: For any $d \in D$, $\Delta(d)$ is given by

$$\Delta(d)(s,t) = \vee_{a \in A} H_{\hat{d}}(\delta(s,a), \delta(t,a))$$

for all $s, t \in S$.

There is no difficulty to check that $\Delta(d) \in \mathcal{D}$ by Lemma 3, and thereby, $\Delta$ is well-defined.

Our next objective is to show that $\Delta$ has a greatest fixed point with respect to the partial order in Definition 4. Recall that the remarkable Tarski's theorem says that each monotonic function on a complete lattice has a greatest fixed point [31]. Therefore, to show that $\Delta$ has a greatest fixed point, it remains to verify that $\Delta$ is monotonic with respect to $\preceq$.

Recall that for a partially ordered set $(X, \leq)$, a function $f : X \longrightarrow X$ is said to be *monotonic* if for all $x_1, x_2 \in X$, $x_1 \leq x_2$ implies that $f(x_1) \leq f(x_2)$.

*Lemma 4:* The function $\Delta : \mathcal{D} \longrightarrow \mathcal{D}$ is monotonic with respect to the partial order $\preceq$.

*Proof:* See Appendix A. ∎

### C. Behavioral Distance

Based on the results obtained in the previous subsections, we can now define the behavioral distance and present some interesting properties.

Recall that for any function $f : X \longrightarrow X$, an element $x \in X$ is called a *fixed point* of $f$ if $x = f(x)$. By using the lemmas established above, we get a fixed point of $\Delta$.

*Theorem 2:* The function $\Delta : \mathcal{D} \longrightarrow \mathcal{D}$ has a greatest fixed point given by

$$\Delta_{\max} = \sqcup \{d \in \mathcal{D} : d \preceq \Delta(d)\}.$$

*Proof:* Both the existence and explicit representation of the greatest fixed point follow immediately from Lemmas 2 and 4, and Tarski's fixed point theorem. ∎

The greatest fixed point $\Delta_{\max}$ is a pseudo-ultrametric on $S$, which serves as a distance measure on the state set of an FTS.

*Definition 7:* Let $(S, A, \delta)$ be an FTS. For any $s, t \in S$, the *behavioral distance* between $s$ and $t$, denoted $d_f(s,t)$, is defined as

$$d_f(s,t) = \Delta_{\max}(s,t).$$

Due to the ultrametricity of $d_f$, we can make a useful observation.

*Corollary 1:* For any given $\lambda \in [0,1]$, define $R_\lambda = \{(s,t) \in S \times S : d_f(s,t) \leq \lambda\}$. Then $R_\lambda$ is an equivalence relation on $S$.

*Proof:* It follows directly from the three properties of pseudo-ultrametric. ∎

The significance of this corollary is that setting a threshold $\lambda$, which depends on the particular application considered, one can partition the state space such that any two states in the same block have distance at most $\lambda$. We defer describing $R_0$ in the next section.

As an immediate consequence of Theorem 2, we get a way to calculate the behavioral distance for image-finite FTSs. Here, by image-finiteness we mean that for any $s \in S$ and $a \in A$, the cardinality of $\delta(s, a)$ is finite.

*Corollary 2:* Let $(S, A, \delta)$ be an image-finite FTS. Define $\Delta^0(\top) = \top$ and $\Delta^{n+1}(\top) = \Delta(\Delta^n(\top))$, where $\top$ is given by $\top(s, t) = 0$ for all $s, t \in S$. Then
$$d_f = \Delta_{\max} = \sqcap \{\Delta^n(\top) : n \in \mathbb{N}\}.$$

*Proof:* See Appendix A. ∎

For the sake of illustrating the above notions and results, we give a simple example.

*Example 2:* Consider the FTS shown in Fig. 1, where the sates are in circles and each fuzzy transition is depicted via two parts: an arrow for nondeterministic choice (for simplicity, in this example we associate to each state at most one fuzzy transition) and a bunch of arrows for the possibility degrees of entering next states. Formally, $S = \{s_1, s_2, s_3, s_4\}$, $A = \{a\}$, and $\delta$ is defined as follows:
$$\delta(s_1, a) = \{\mu\}, \quad \delta(s_2, a) = \{\eta\},$$
$$\delta(s_3, a) = \{\theta\}, \quad \delta(s_4, a) = \emptyset,$$
where
$$\mu = \frac{0.9}{s_3} + \frac{0.8}{s_4}, \quad \eta = \frac{0.6}{s_3} + \frac{0.9}{s_4}, \text{ and } \theta = \frac{0.9}{s_4}.$$

Using Corollary 2, we compute $d_f(s_i, s_j)$ by iteration of $\Delta$ starting from the greatest element $\top$. Let us write $d_n$ for $\Delta^n(\top)$. By the properties (P1) and (P2) of pseudo-ultrametric, we only need to calculate $d_n(s_i, s_j)$ with $i < j$. Moreover, it follows by the monotony of $\Delta$ that $d_{n+1}(s_i, s_j) = 1$ if $d_n(s_i, s_j) = 1$. These observations greatly simplify the computation below. By Definitions 3 and 5, we have the following:
$$\begin{aligned}
d_1(s_1, s_2) &= \Delta(\top)(s_1, s_2) = H_{\hat{\top}}(\delta(s_1, a), \delta(s_2, a)) \\
&= H_{\hat{\top}}(\{\mu\}, \{\eta\}) = \hat{\top}(\mu, \eta) = 0, \\
d_1(s_1, s_3) &= \hat{\top}(\mu, \theta) = 0, \\
d_1(s_1, s_4) &= \hat{\top}(\mu, \emptyset) = 1, \\
d_1(s_2, s_3) &= \hat{\top}(\eta, \theta) = 0, \\
d_1(s_2, s_4) &= \hat{\top}(\eta, \emptyset) = 1, \\
d_1(s_3, s_4) &= \hat{\top}(\theta, \emptyset) = 1.
\end{aligned}$$

For clarity, we may represent $d_1$ by the matrix $(d_1(s_i, s_j))$, i.e.,
$$d_1 = \begin{bmatrix} 0 & 0 & 0 & 1 \\ 0 & 0 & 0 & 1 \\ 0 & 0 & 0 & 1 \\ 1 & 1 & 1 & 0 \end{bmatrix}$$

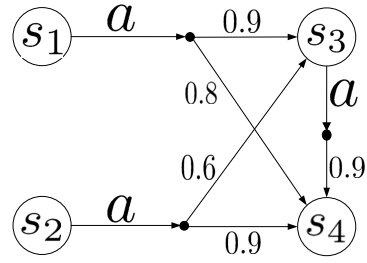

Fig. 1. An FTS.

Let us go ahead.
$$\begin{aligned}
d_2(s_1, s_2) &= \Delta(d_1)(s_1, s_2) = H_{\hat{d}_1}(\delta(s_1, a), \delta(s_2, a)) \\
&= H_{\hat{d}_1}(\{\mu\}, \{\eta\}) = \hat{d}_1(\mu, \eta) = 0.9, \\
d_2(s_1, s_3) &= \hat{d}_1(\mu, \theta) = 0.9, \\
d_2(s_2, s_3) &= \hat{d}_1(\eta, \theta) = 0.6.
\end{aligned}$$

Again, we represent $d_2$ by the matrix
$$d_2 = \begin{bmatrix} 0 & 0.9 & 0.9 & 1 \\ 0.9 & 0 & 0.6 & 1 \\ 0.9 & 0.6 & 0 & 1 \\ 1 & 1 & 1 & 0 \end{bmatrix}$$

We need to proceed with iteration. Fortunately, the next step shows us that $d_3 = d_2$ and thus the iteration can be stopped at the third step. As a result, we obtain that $d_f = d_2$, that is,
$$\begin{aligned}
d_f(s_i, s_i) &= 0, \text{ for all } i = 1, 2, 3, 4, \\
d_f(s_2, s_3) &= 0.6, \\
d_f(s_1, s_2) &= d_f(s_1, s_3) = 0.9, \\
d_f(s_1, s_4) &= d_f(s_2, s_4) = d_f(s_3, s_4) = 1,
\end{aligned}$$
where the symmetric parts are elided.

We end this section by relating the behavioral distance with the similarity relation (also known as fuzzy equivalence relation) proposed by Zadeh in [32]. Recall that a *similarity relation* on $X$ is a binary fuzzy relation $\mathcal{S}$ on $X$ (i.e., a function from $X \times X \longrightarrow [0, 1]$) that satisfies $\mathcal{S}(x, x) = 1$, $\mathcal{S}(x, y) = \mathcal{S}(y, x)$, and $\mathcal{S}(x, y) \wedge \mathcal{S}(y, z) \leq \mathcal{S}(x, z)$, for any $x, y, z \in X$. We draw the reader's attention to the entire difference between Zadeh's similarity relation and the notion of bisimilarity in the subsequent section.

The following observation implies that the less the value of $d_f$, the more similar the two states.

*Corollary 3:* For any $s, t \in S$, let $\mathcal{S}(s, t) = 1 - d_f(s, t)$. Then $\mathcal{S}$ is a similarity relation on $S$.

*Proof:* It follows immediately from the fact that $d_f$ is a pseudo-ultrametric on $S$. ∎

## IV. RELATIONSHIP WITH BISIMILARITY

In this section, we embark upon the relationship between the behavioral distance and bisimilarity. More concretely, we will show that two states are bisimilar if and only if they have behavioral distance 0. In addition, we present a characterization of bisimulation by exploiting the function $\Delta$. To this end, we need to lift a relation on states to a relation on possibility distributions.

## A. Lifting of Relation

The following notion of lifting is adopted from [33], where a similar notion was first defined for probability distributions.

*Definition 8:* Given an $R \subseteq S \times S$, the lifting $\hat{R}$ of $R$ is defined as the smallest binary relation on $\mathcal{F}(S)$ that satisfies
(1) $(s,t) \in R$ implies $(\hat{s}, \hat{t}) \in \hat{R}$;
(2) $(\mu_i, \eta_i) \in \hat{R}$ implies $(\bigcup_{i \in I} c_i \cdot \mu_i, \bigcup_{i \in I} c_i \cdot \eta_i) \in \hat{R}$, for any finite index set $I$ and $c_i \in [0,1]$.

Note that $\hat{s}$ in the above definition stands for the possibility distribution $\frac{1}{s}$.

For later need, it is convenient to have two alternative presentations of the lifting.

*Lemma 5:* Let $\mu, \eta \in \mathcal{F}(S)$ and $R \subseteq S \times S$. Then the following are equivalent.
(1) $(\mu, \eta) \in \hat{R}$.
(2) There exist $(s_i, t_i) \in R$ and $c_i \in [0,1]$, $i \in I$, such that $\mu = \bigcup_{i \in I} c_i \cdot \hat{s}_i$ and $\eta = \bigcup_{i \in I} c_i \cdot \hat{t}_i$.
(3) There is a weight function $w: S \times S \longrightarrow [0,1]$ such that
(a) $\vee_{t \in S} w(s,t) = \mu(s)$ for any $s \in S$;
(b) $\vee_{s \in S} w(s,t) = \eta(t)$ for any $t \in S$;
(c) $w(s,t) > 0$ implies $(s,t) \in R$.

*Proof:* See Appendix A. ∎

For the special case of an equivalence relation, there is a simpler way to describe the lifting.

*Lemma 6:* Suppose that $\mu, \eta \in \mathcal{F}(S)$ and $R$ is an equivalence relation on $S$. Then $(\mu, \eta) \in \hat{R}$ if and only if $\mu(C) = \eta(C)$ for all $C \in S/R$.

*Proof:* See Appendix A. ∎

As we have seen, Definition 2 lifts a pseudo-ultrametric $d$ on $S$ to a pseudo-ultrametric $\hat{d}$ on $\mathcal{F}(S)$, while Definition 8 lifts a binary relation $R$ on $S$ to a binary relation $\hat{R}$ on $\mathcal{F}(S)$. With a little surprise, there is an intrinsic connection between them.

*Lemma 7:* Suppose that $R$ is a binary relation and $d$ is a pseudo-ultrametric on $S$ satisfying that for any $s,t \in S$,

$$(s,t) \in R \text{ if and only if } d(s,t) = 0. \quad (2)$$

Then it holds that for any $\mu, \eta \in \mathcal{F}(S)$,

$$(\mu, \eta) \in \hat{R} \text{ if and only if } \hat{d}(\mu, \eta) = 0. \quad (3)$$

*Proof:* See Appendix A. ∎

In order to characterize bisimulation via the function $\Delta$, we need to define an auxiliary function. For any relation $R \subseteq S \times S$, we associate to it a function $d_R: S \times S \longrightarrow [0,1]$ defined by

$$d_R(s,t) = \begin{cases} 0, & \text{if } (s,t) \in R \\ 1, & \text{otherwise.} \end{cases}$$

Let us make a useful observation on which sort relation makes $d_R$ into a pseudo-ultrametric.

*Lemma 8:* Let $R \subseteq S \times S$. Then $R$ is an equivalence relation if and only if $d_R$ is a pseudo-ultrametric.

*Proof:* It is straightforward. We thus omit the details. ∎

## B. Relationship

Let us begin with the notion of bisimulation from [8], which is the nondeterministic version of the bisimulation in [7].

*Definition 9:* Let $(S, A, \delta)$ be an FTS. An equivalence relation $R \subseteq S \times S$ is called a *bisimulation* on $S$ if for any $(s,t) \in R$ and $a \in A$, $s \xrightarrow{a} \mu$ implies $t \xrightarrow{a} \eta$ for some $\eta$ such that $\mu(C) = \eta(C)$ for every $C \in S/R$.

The greatest bisimulation on $S$, denoted by $\sim$, is called *bisimilarity*. In other words, $s$ and $t$ are *bisimilar*, written $s \sim t$, if $(s,t) \in R$ for some bisimulation $R$.

It is easy to see that in Fig. 1, $s_i \not\sim s_j$ if $i \neq j$.

We will characterize the bisimulation by the post-fixed points of $\Delta$. Recall that for a partially ordered set $(X, \leq)$ and a function $f: X \longrightarrow X$, an element $x$ of $X$ is called a *post-fixed point* of $f$ if $x \leq f(x)$.

For the proofs of subsequent theorems, let us present an explicit characterization of the post-fixed points of $\Delta$.

*Lemma 9:* For any $d \in \mathcal{D}$, $d$ is a post-fixed point of $\Delta$ if and only if for any $(s,t) \in S \times S$ and $a \in A$, $s \xrightarrow{a} \mu$ implies that there exists some $\eta$ (possibly $\emptyset$) such that $t \xrightarrow{a} \eta$ and $\hat{d}(\mu, \eta) \leq d(s,t)$.

*Proof:* Note that $d$ is a post-fixed point of $\Delta$ if and only if $\vee_{a \in A} H_{\hat{d}}(\delta(s,a), \delta(t,a)) \leq d(s,t)$ for all $(s,t) \in S \times S$. The remainder of the proof follows readily from the definition of Hausdorff distance. ∎

We can now state the first main theorem in this section.

*Theorem 3:* Let $R$ be an equivalence relation on $S$. Then $R$ is a bisimulation if and only if $d_R$ is a post-fixed point of $\Delta$.

*Proof:* See Appendix A. ∎

The next theorem characterizing bisimilarity justifies the soundness of our behavioral distance.

*Theorem 4:* For any $s,t \in S$, $s \sim t$ if and only if $d_f(s,t) = 0$.

*Proof:* See Appendix A. ∎

Following the notation used in Corollary 1, the above theorem means exactly that $R_0 = \sim$. As an immediate consequence of the theorem, we have the following.

*Corollary 4:* For any $s, s', t, t' \in S$, if $s \sim s'$ and $t \sim t'$, then $d_f(s,t) = d_f(s',t')$.

*Proof:* See Appendix A. ∎

It is well known that bisimulation equivalence allows one to reduce the state space of a system by combining bisimilar states to generate a quotient system with an equivalent behavior but with fewer states. Therefore, the significance of Corollary 4 lies in that computing the behavioral distance between $s$ and $t$ is reduced to computing the behavioral distance between their quotients.

## V. NON-EXPANSIVENESS

To construct an overall fuzzy system, one may usually build its component systems first and then compose them by some operators. Therefore, compositional operators can serve the need of modular specification and verification of systems. A desirable property of compositional operators is non-expansiveness. It means that if the difference (with respect to some behavioral measure) between $s_i$ and $s'_i$ is $\epsilon_i$, then the difference between $f(s_1,\ldots,s_n)$ and $f(s'_1,\ldots,s'_n)$ is no more than $\vee_{i=1}^n \epsilon_i$, where $f$ is an operator with $n$ arguments. As an example, we examine the non-expansiveness of our behavioral distance with respect to the parallel composition and product operators in this section. These operators model two forms of joint behavior of some fuzzy systems and we can think of them as two types of systems resulting from the interconnection of system components.

To introduce the operators, one further bit of notation will be handy: Given $\mu_i \in \mathcal{F}(S_i)$, $i = 1, 2$, we use $\mu_1 \wedge \mu_2$ to denote the possibility distribution on $S_1 \times S_2$ that is defined by

$$(\mu_1 \wedge \mu_2)(s_1, s_2) = \mu_1(s_1) \wedge \mu_2(s_2)$$

for all $(s_1, s_2) \in S_1 \times S_2$.

The parallel composition operator is asynchronous in the sense that the components can either synchronize or act independently. Given an FTS $(S, A, \delta)$, for any $s_1, s_2 \in S$, the events that are intended to synchronize at $s_1$ and $s_2$ are listed in the set $A_{s_1} \cap A_{s_2}$ and the rest of the events can be performed independently, where $A_{s_i}$ stands for $\{a \in A : \exists \mu \neq \emptyset, s_i \xrightarrow{a} \mu\}$. Formally, the *parallel composition* is a three-tuple $(S \times S, A, \delta')$, where for all $s_1, s_2 \in S$ and $a \in A$,

$\delta'(s_1|s_2, a) =$

$$\begin{cases} \{\mu \wedge \hat{q} : \mu \in \delta(s_1, a)\}, & \text{if } a \in A_{s_1} \backslash A_{s_2} \\ \{\hat{p} \wedge \eta : \eta \in \delta(s_2, a)\}, & \text{if } a \in A_{s_2} \backslash A_{s_1} \\ \{\mu \wedge \eta : \mu \in \delta(s_1, a), \eta \in \delta(s_2, a)\}, & \text{if } a \in A_{s_1} \cap A_{s_2} \\ \emptyset, & \text{otherwise.} \end{cases}$$

Clearly, this constructs an FTS, which represents that two states of $(S, A, \delta)$ are running concurrently. Instead of a pair $(s_1, s_2) \in S \times S$ we write $s_1|s_2$ for a state in the composed FTS. The synchronization constraint $A_{s_1} \cap A_{s_2}$ forces some events to be carried out at both of the states at the same time and allows all the possible interleavings of the other events at the two states.

The definition of product is somewhat simpler than that of the parallel composition. Given an FTS $(S, A, \delta)$, the *product* is a three-tuple $(S \times S, A, \delta'')$, where $\delta'' : (S \times S) \times A \longrightarrow \mathcal{P}(\mathcal{F}(S \times S))$ is defined by

$\delta''(s_1 \| s_2, a) =$

$$\begin{cases} \{\mu \wedge \eta : \mu \in \delta(s_1, a), \eta \in \delta(s_2, a)\}, & \text{if } a \in A_{s_1} \cap A_{s_2} \\ \emptyset, & \text{otherwise.} \end{cases}$$

It turns out that $(S \times S, A, \delta'')$ is again an FTS. The product requires that the components are strictly synchronous.

Let us present an example of the parallel composition and compute the behavioral distance between the states.

*Example 3:* We revisit the FTS shown in Fig. 1. By the definition of parallel composition, we can readily get all the fuzzy transitions derived from $s_1|s_2$ and $s_2|s_3$, which are depicted in Fig. 2. For instance,

$$\begin{aligned} \delta'(s_2|s_3, a) &= \left\{ \left(\frac{0.6}{s_3} + \frac{0.9}{s_4}\right) \wedge \frac{0.9}{s_4} \right\} \\ &= \left\{ \frac{0.6}{s_3|s_4} + \frac{0.9}{s_4|s_4} \right\}, \\ \delta'(s_3|s_4, a) &= \left\{ \frac{0.9}{s_4} \wedge \frac{1}{s_4} \right\} \\ &= \left\{ \frac{0.9}{s_4|s_4} \right\}. \end{aligned}$$

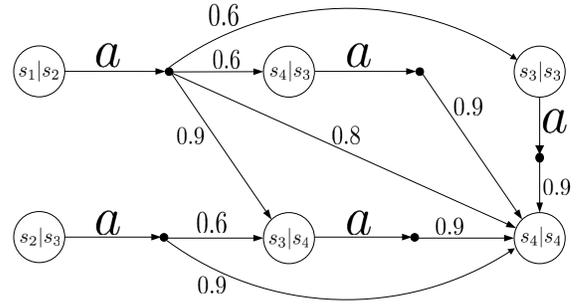

Fig. 2. The partial transition graph of parallel composition derived from $s_1|s_2$ and $s_2|s_3$.

By a routine computation like Example 2, we can obtain the values of behavioral distance $d'_f$ on these states; for example,

$$\begin{aligned} d'_f(s_4|s_3, s_3|s_4) &= d'_f(s_4|s_3, s_3|s_3) = d'_f(s_3|s_4, s_3|s_3) \\ &= 0, \\ d'_f(s_2|s_3, s_4|s_3) &= d'_f(s_2|s_3, s_3|s_4) = d'_f(s_2|s_3, s_3|s_3) \\ &= 0.6, \\ d'_f(s_1|s_2, s_2|s_3) &= d'_f(s_1|s_2, s_4|s_3) = d'_f(s_1|s_2, s_3|s_4) \\ &= d'_f(s_1|s_2, s_3|s_3) = 0.9, \\ d'_f(s_1|s_2, s_4|s_4) &= d'_f(s_2|s_3, s_4|s_4) = d'_f(s_4|s_3, s_4|s_4) \\ &= d'_f(s_3|s_4, s_4|s_4) = d'_f(s_3|s_3, s_4|s_4) = 1. \end{aligned}$$

Comparing with the behavioral distances obtained in Example 2, these values evidence that the parallel composition is non-expansive.

More generally, we have the following theorem.

*Theorem 5:* Let $(S, A, \delta)$ be an FTS and $s_i, t_i \in S$ with $d_f(s_i, t_i) = \epsilon_i$, where $i = 1, 2$. Then we have the following:
(1) $d'_f(s_1|s_2, t_1|t_2) \leq \epsilon_1 \vee \epsilon_2$.
(2) $d''_f(s_1\|s_2, t_1\|t_2) \leq \epsilon_1 \vee \epsilon_2$.

*Proof:* See Appendix A. ∎

The above theorem gives rise to an interesting corollary, which says that both the parallel composition and product preserve bisimilarity.

*Corollary 5:* If $s_1 \sim t_1$ and $s_2 \sim t_2$, then $s_1|s_2 \sim t_1|t_2$ and $s_1\|s_2 \sim t_1\|t_2$.

*Proof:* It is straightforward by Theorems 4 and 5. ∎

## VI. CONCLUSION AND FUTURE WORK

In this paper, we have constructed a pseudo-ultrametric for measuring the behavioral distance between states in an FTS. The behavioral distance is a quantitative analogue of bisimilarity. The smaller the distance, the more alike the states are. In particular, states are bisimilar if and only if they have a distance of 0. We have also shown that for the parallel composition and product operators, the behavioral distance is non-expansive. Contrary to the exact notion of bisimulation, the behavioral distance is much more natural and robust for fuzzy systems, which enables us to talk about the approximate equivalence of fuzzy systems.

There are several problems that are worth further study. First, developing an algorithm to calculate our behavioral distance is desirable. The algorithms for probabilistic systems cannot be adapted in a straightforward way to our setting. Second, another aspect that should be considered in the future is to modify the logic and its interpretation in [34], [35] so that we can obtain a logical characterization of the behavioral distance. This is just a different formal presentation of the same thing, but it may help in connecting to the fairly large number of works on fuzzy logic. Third, it is interesting to discover the relationship between the behavioral distance and the greatest bisimulation based on a fuzzy relation in [12], [13]. It would be very valuable if one could transfer results from one setting to the other. Fourth, noting that there is another approach to measuring the behavioral distance between states in a probabilistic system by introducing various approximate bisimulations (e.g., [16], [36]–[39]), one may pursue the approximate bisimulation approach in a fuzzy system as well. Finally, the present work has focused on the theoretical presentation of our behavioral distance. The application of this measure in analyzing some practical fuzzy systems such as fuzzy control systems, discrete-time fuzzy systems [40], and formal models of computing with words [41]–[43] is left for future work.

## APPENDIX A

*Proof of Lemma 1:* Suppose that the system (1) has a solution $\{x_{st} : s, t \in S\}$. By contradiction, let us assume that $\mu(S) \neq \eta(S)$, say $\mu(S) > \eta(S)$. Then there exists $s' \in S$ such that $\mu(s') = \mu(S)$. Because $\mu(s') = \vee_{t \in S} x_{s't}$, there is $t' \in S$ such that $x_{s't'} = \mu(s')$. Hence, we see that $x_{s't'} > \eta(S) \geq \eta(t')$. Furthermore, it yields that $\vee_{s \in S} x_{st'} \geq x_{s't'} > \eta(t')$, which contradicts with the second equation in the system. Therefore, the necessity holds.

Conversely, assume that $\mu(S) = \eta(S)$. Then there are $s', t' \in S$ such that $\mu(s') = \mu(S) = \eta(S) = \eta(t')$. For any $s, t \in S$, we take

$$x_{st} = \begin{cases} \mu(s), & \text{if } t = t' \\ \eta(t), & \text{if } s = s' \\ 0, & \text{otherwise.} \end{cases}$$

Note that $x_{s't'} = \mu(s') = \eta(t')$ by the definition above. We claim that $\{x_{st} : s, t \in S\}$ is a solution to the system (1). In fact, it is clear that $x_{st} \geq 0$ for all $s, t \in S$. The first two equations in (1) are analogous, so we merely verify the first one. For any $s \in S$, if $s = s'$, then we have that $\vee_{t \in S} x_{s't} = \vee_{t \in S} \eta(t) = \eta(S) = \mu(S) = \mu(s')$; if $s \neq s'$, we get that $\vee_{t \in S} x_{st} = x_{st'} \vee (\vee_{t \in S \setminus \{t'\}} x_{st}) = \mu(s) \vee 0 = \mu(s)$.

Consequently, it holds that $\vee_{t \in S} x_{st} = \mu(s)$ for any $s \in S$, as desired. ∎

*Proof of Theorem 1:* We need to check that $\hat{d}$ satisfies the three requirements of pseudo-ultrametric. Let $\mu, \eta, \theta \in \mathcal{F}(S)$.

For (P1), we see that $\hat{d}(\mu, \mu) = 0$ by setting $x_{ss} = \mu(s)$ and $x_{st} = 0$ for all $s \neq t$.

For (P2), if $\mu(S) \neq \eta(S)$, we have that $\hat{d}(\mu, \eta) = 1 = \hat{d}(\eta, \mu)$. Otherwise, without loss of generality, we may assume that $\hat{d}(\mu, \eta) < \hat{d}(\eta, \mu)$, and suppose that $\hat{d}(\mu, \eta)$ is attained by some $\{x_{st} : s, t \in S\}$, namely, $\hat{d}(\mu, \eta) = \vee_{s,t \in S} (d(s,t) \wedge x_{st})$. For all $s, t \in S$, set $y_{st} = x_{ts}$. It is easy to see that $\vee_{t \in S} y_{st} = \vee_{t \in S} x_{ts} = \eta(s)$ for every $s \in S$ and $\vee_{s \in S} y_{st} = \vee_{s \in S} x_{ts} = \mu(t)$ for every $t \in S$. Moreover, we obtain that

$$\begin{aligned} \vee_{s,t \in S} (d(s,t) \wedge y_{st}) &= \vee_{s,t \in S} (d(s,t) \wedge x_{ts}) \\ &= \vee_{s,t \in S} (d(t,s) \wedge x_{ts}) \\ &= \vee_{s,t \in S} (d(s,t) \wedge x_{st}) \\ &= \hat{d}(\mu, \eta), \end{aligned}$$

which means that $\hat{d}(\mu, \eta) \geq \hat{d}(\eta, \mu)$. It is a contradiction. Therefore, $\hat{d}(\mu, \eta) = \hat{d}(\eta, \mu)$.

For (P3), if $\mu(S) \neq \eta(S)$ or $\eta(S) \neq \theta(S)$, we always have that $\hat{d}(\mu, \eta) \vee \hat{d}(\eta, \theta) = 1 \geq \hat{d}(\mu, \theta)$. Otherwise, we see that $\mu(S) = \eta(S) = \theta(S)$. In this case, suppose that $\hat{d}(\mu, \eta) = \vee_{s,t \in S} (d(s,t) \wedge x_{st})$ and $\hat{d}(\eta, \theta) = \vee_{s,t \in S} (d(s,t) \wedge y_{st})$. Then we have the following equations:

$$\vee_{t \in S} x_{st} = \mu(s), \quad \forall s \in S \quad (4)$$
$$\vee_{s \in S} x_{st} = \eta(t), \quad \forall t \in S \quad (5)$$
$$\vee_{t \in S} y_{st} = \eta(s), \quad \forall s \in S \quad (6)$$
$$\vee_{s \in S} y_{st} = \theta(t), \quad \forall t \in S \quad (7)$$

For all $s, t \in S$, set $z_{st} = \vee_{r \in S} (x_{sr} \wedge y_{rt})$. For any given $s \in S$, we get that

$$\begin{aligned} \vee_{t \in S} z_{st} &= \vee_{t \in S} \vee_{r \in S} (x_{sr} \wedge y_{rt}) \\ &= \vee_{r \in S} \vee_{t \in S} (x_{sr} \wedge y_{rt}) \\ &= \vee_{r \in S} [x_{sr} \wedge (\vee_{t \in S} y_{rt})] \\ &\stackrel{(6)}{=} \vee_{r \in S} [x_{sr} \wedge \eta(r)] \\ &\stackrel{(5)}{=} \vee_{r \in S} x_{sr} \\ &\stackrel{(4)}{=} \mu(s), \end{aligned}$$

namely, $\vee_{t \in S} z_{st} = \mu(s)$. At the same time, for any $t \in S$, we have that

$$\begin{aligned} \vee_{s \in S} z_{st} &= \vee_{s \in S} \vee_{r \in S} (x_{sr} \wedge y_{rt}) \\ &= \vee_{r \in S} \vee_{s \in S} (x_{sr} \wedge y_{rt}) \\ &= \vee_{r \in S} [(\vee_{s \in S} x_{sr}) \wedge y_{rt}] \\ &\stackrel{(5)}{=} \vee_{r \in S} [\eta(r) \wedge y_{rt}] \\ &\stackrel{(6)}{=} \vee_{r \in S} y_{rt} \\ &\stackrel{(7)}{=} \theta(t), \end{aligned}$$

i.e., $\vee_{s \in S} z_{st} = \theta(t)$. Moreover, we see that $\hat{d}(\mu, \theta) \leq \hat{d}(\mu, \eta) \vee \hat{d}(\eta, \theta)$ by the inequality at the top of the next page.

$$\begin{aligned}
\hat{d}(\mu,\theta) &\leq \vee_{s,t\in S}(d(s,t)\wedge z_{st}) \\
&= \vee_{s,t\in S}[d(s,t)\wedge(\vee_{r\in S}(x_{sr}\wedge y_{rt}))] \\
&= \vee_{s,t\in S}\vee_{r\in S}[d(s,t)\wedge x_{sr}\wedge y_{rt}] \\
&\leq \vee_{s,t\in S}\vee_{r\in S}[(d(s,r)\vee d(r,t))\wedge x_{sr}\wedge y_{rt}] \\
&= \vee_{s,t\in S}\vee_{r\in S}[(d(s,r)\wedge x_{sr}\wedge y_{rt})\vee(d(r,t)\wedge x_{sr}\wedge y_{rt})] \\
&\leq \vee_{s,t\in S}\vee_{r\in S}[(d(s,r)\wedge x_{sr})\vee(d(r,t)\wedge y_{rt})] \\
&= [\vee_{s,t\in S}\vee_{r\in S}(d(s,r)\wedge x_{sr})]\vee[\vee_{s,t\in S}\vee_{r\in S}(d(r,t)\wedge y_{rt})] \\
&= [\vee_{s,r\in S}(d(s,r)\wedge x_{sr})]\vee[\vee_{r,t\in S}(d(r,t)\wedge y_{rt})] \\
&= \hat{d}(\mu,\eta)\vee\hat{d}(\eta,\theta).
\end{aligned}$$

Therefore, $\hat{d}$ is a pseudo-ultrametric on $\mathcal{F}(S)$, completing the proof of the theorem. ∎

*Proof of Lemma 2:* For any $X\subseteq\mathcal{D}$, we define $\sqcap X$ by

$$(\sqcap X)(s,t)=\vee\{d(s,t):d\in X\}$$

and $\sqcup X$ by

$$\sqcup X=\sqcap\{d\in\mathcal{D}:\forall d'\in X, d'\preceq d\}.$$

In particular, $(\sqcap\emptyset)(s,t)=0$ for all $s,t\in S$, and $(\sqcup\emptyset)(s,t)=0$ if $s=t$, 1 otherwise. We need to verify that $\sqcap X$ and $\sqcup X$ are the infimum and supremum of $X$, respectively. We only prove that $\sqcap X$ is the infimum, since the supremum can be proved similarly.

We first show that $\sqcap X\in\mathcal{D}$. In fact, it is obvious that $(\sqcap X)(s,s)=0$ and $(\sqcap X)(s,t)=(\sqcap X)(t,s)$ for all $s,t\in S$. For (P3), we have that

$$\begin{aligned}
(\sqcap X)(s,t) &= \vee_{d\in X}d(s,t) \\
&\leq \vee_{d\in X}(d(s,r)\vee d(r,t)) \\
&= (\vee_{d\in X}d(s,r))\vee(\vee_{d\in X}d(r,t)) \\
&= [(\sqcap X)(s,r)]\vee[(\sqcap X)(r,t)],
\end{aligned}$$

for any $s,t,r\in S$. Hence, $\sqcap X$ is a pseudo-ultrametric.

Observe that $(\sqcap X)(s,t)\geq d(s,t)$, that is, $\sqcap X\preceq d$, for all $d\in X$. Therefore, $\sqcap X$ is a lower bound of $X$. On the other hand, for any $h\in\mathcal{D}$, if $h\preceq d$, i.e., $h(s,t)\geq d(s,t)$, for all $d\in X$, then it is easy to see that $h\preceq\sqcap X$, which means that $\sqcap X$ is not less than any other lower bound of $X$. Consequently, $\sqcap X$ is the infimum of $X$. This proves that $(\mathcal{D},\preceq)$ is a complete lattice. ∎

*Proof of Lemma 4:* We first prove the following claim: For any given $d_1,d_2\in\mathcal{D}$, if $d_1\preceq d_2$, then it holds that $\hat{d}_1(\mu,\eta)\geq\hat{d}_2(\mu,\eta)$ for all $\mu,\eta\in\mathcal{F}(S)$.

In fact, we have that $d_1(s,t)\geq d_2(s,t)$ for all $s,t\in S$, as $d_1\preceq d_2$. Suppose that $\hat{d}_1(\mu,\eta)$ is achieved by $\{x_{st}:s,t\in S\}$. Clearly, $\{x_{st}:s,t\in S\}$ is also a feasible solution to (MP) defining $\hat{d}_2(\mu,\eta)$. As a result, we obtain that

$$\begin{aligned}
\hat{d}_1(\mu,\eta) &= \vee_{s,t\in S}(d_1(s,t)\wedge x_{st}) \\
&\geq \vee_{s,t\in S}(d_2(s,t)\wedge x_{st}) \\
&\geq \hat{d}_2(\mu,\eta),
\end{aligned}$$

as desired.

The proof of the lemma is straightforward by the previous claim and thus omitted here. ∎

*Proof of Corollary 2:* According to Tarski's fixed point theorem, the greatest fixed point $\Delta_{\max}$ can be obtained by iteration of $\Delta$ starting from the greatest element $\top$ (see, for example, [44]). Hence, to show that $\Delta_{\max}=\sqcap\{\Delta^n(\top):n\in\mathbb{N}\}$, we only need to prove that the closure ordinal of $\Delta$, i.e., the least ordinal $n$ such that $\Delta^{n+1}=\Delta^n$, is at most $\omega$. In fact, for any $(s,t)\in S\times S$, if $s\xrightarrow{a}\mu$, then for each $d_n=\Delta^n(\top)$, there exists $\eta_n$ such that $t\xrightarrow{a}\eta_n$ and $\hat{d}_n(\mu,\eta_n)\leq d_n(s,t)$. Thanks to the image-finiteness of the FTS, there is an $\eta_n$, say $\eta$, such that $t\xrightarrow{a}\eta$ and $\hat{d}_n(\mu,\eta)\leq d_n(s,t)$ for all but finitely many $n$, as desired. ∎

*Proof of Lemma 5:* The equivalence of (1) and (2) follows immediately from Definition 8. Therefore, we only prove the equivalence of (2) and (3).

(2) $\Rightarrow$ (3). By the condition of (2), we define the weight function $w$ by setting

$$w(s,t)=\vee\{c_i:(s_i,t_i)=(s,t),i\in I\}$$

for each $(s,t)\in S\times S$. It remains to check that such a weight function satisfies the requirements (a), (b), and (c) of (3).

(a) For any $s\in S$, we have that

$$\begin{aligned}
\vee_{t\in S}w(s,t) &= \vee_{t\in S}\vee\{c_i:(s_i,t_i)=(s,t),i\in I\} \\
&= \vee\{c_i:s_i=s,i\in I\} \\
&= \vee_{i\in I}[c_i\wedge\hat{s}_i(s)] \\
&= \mu(s).
\end{aligned}$$

(b) The proof is similar to that of $(a)$.

(c) If $w(s,t)>0$, then by the definition of $w$ there exists some $i\in I$ such that $c_i>0$ and $(s_i,t_i)=(s,t)$. We see that $(s,t)\in R$, since $(s_i,t_i)\in R$.

(3) $\Rightarrow$ (2). Suppose that there is a weight function $w$ satisfying the conditions (a), (b), and (c) in (3). Then we take $I=\{(s,t):w(s,t)>0,(s,t)\in S\times S\}$ and set $c_{(s,t)}=w(s,t)$ for each $(s,t)\in I$. Clearly, for any $(s,t)\in I$, we have that $w(s,t)>0$, which implies $(s,t)\in R$ by the

condition (c) in (3). Moreover, for any $s' \in S$, we obtain that

$$\Big(\bigcup_{(s,t)\in I} c_{(s,t)} \cdot \hat{s}\Big)(s') = \vee_{(s,t)\in I}(w(s,t) \wedge \hat{s}(s'))$$
$$= \vee_{(s',t)\in I} w(s',t)$$
$$= \vee\{w(s',t) : w(s',t) > 0, t \in S\}$$
$$= \vee\{w(s',t) : t \in S\}$$
$$= \mu(s'),$$

which means that $\mu = \bigcup_{(s,t)\in I} c_{(s,t)} \cdot \hat{s}$. By the same token, we can get that $\eta = \bigcup_{(s,t)\in I} c_{(s,t)} \cdot \hat{t}$. Hence, (2) holds. This completes the proof of the lemma. ∎

*Proof of Lemma 6:* We first show the necessity. Suppose that $(\mu, \eta) \in \hat{R}$. Then by Lemma 5, there are $(s_i, t_i) \in R$ and $c_i \in [0,1]$, $i \in I$, such that $\mu = \bigcup_{i\in I} c_i \cdot \hat{s}_i$ and $\eta = \bigcup_{i\in I} c_i \cdot \hat{t}_i$. For any $C \in S/R$, we have that

$$\mu(C) = \vee_{s\in C} \mu(s)$$
$$= \vee_{s\in C} \Big(\bigcup_{i\in I} c_i \cdot \hat{s}_i\Big)(s)$$
$$= \vee_{s\in C} \vee \{c_i : i \in I, s_i = s\}$$
$$= \vee\{c_i : i \in I, s_i \in C\}.$$

In a similar vein, we can get that $\eta(C) = \vee\{c_i : i \in I, t_i \in C\}$. As $(s_i, t_i) \in R$, we see that $s_i \in C$ if and only if $t_i \in C$. Consequently, it yields that $\mu(C) = \eta(C)$.

Next, to see the sufficiency, assume that $\mu(C) = \eta(C)$ for every $C \in S/R$. Take $I = \{(s,t) : (s,t) \in R\}$ and let $c_{(s,t)} = \mu(s) \wedge \eta(t)$ for each $(s,t) \in I$. For any $(s,t) \in I$, it is obvious that $(s,t) \in R$. Furthermore, for any $s' \in S$, we have that

$$\Big(\bigcup_{(s,t)\in I} c_{(s,t)} \cdot \hat{s}\Big)(s')$$
$$= \vee_{(s,t)\in I}(c_{(s,t)} \wedge \hat{s}(s'))$$
$$= \vee_{(s',t)\in I} c_{(s',t)}$$
$$= \vee\{\mu(s') \wedge \eta(t) : (s',t) \in R, t \in S\}$$
$$= \vee\{\mu(s') \wedge \eta(t) : t \in [s']\}$$
$$= \mu(s') \wedge (\vee_{t\in[s']} \eta(t))$$
$$= \mu(s') \wedge \eta([s'])$$
$$= \mu(s') \wedge \mu([s'])$$
$$= \mu(s'),$$

where we write $[s']$ for the equivalence class containing $s'$. It gives that $\mu = \bigcup_{(s,t)\in I} c_{(s,t)} \cdot \hat{s}$. Similarly, we can get that $\eta = \bigcup_{(s,t)\in I} c_{(s,t)} \cdot \hat{t}$. Whence, we have that $(\mu, \eta) \in \hat{R}$ by Lemma 5, finishing the proof. ∎

*Proof of Lemma 7:* First we prove the sufficiency by constructing a weight function $w : S \times S \longrightarrow [0,1]$ that satisfies the requirements (a), (b), and (c) of Lemma 5. Suppose that $\hat{d}(\mu, \eta) = 0$ and it is attained by $\{x_{st} : s, t \in S\}$. It means that $\vee_{s,t\in S}(d(s,t) \wedge x_{st}) = 0$, $\vee_{t\in S} x_{st} = \mu(s)$ for any $s \in S$, and $\vee_{s\in S} x_{st} = \eta(t)$ for any $t \in S$. Simply taking $w(s,t) = x_{st}$, we see that the requirements (a) and (b) of Lemma 5 are satisfied. If $w(s,t) = x_{st} > 0$, then it forces by the argument $\vee_{s,t\in S}(d(s,t) \wedge x_{st}) = 0$ that $d(s,t) = 0$, which implies that $(s,t) \in R$ by (2). It thus follows by Lemma 5 that $(\mu, \eta) \in \hat{R}$.

Now, let us show the necessity. Suppose that $(\mu, \eta) \in \hat{R}$. Then by Lemma 5 there is a weight function $w : S \times S \longrightarrow [0,1]$ such that $\vee_{t\in S} w(s,t) = \mu(s)$ for any $s \in S$ and $\vee_{s\in S} w(s,t) = \eta(t)$ for any $t \in S$. Moreover, $w(s,t) > 0$ implies $(s,t) \in R$, which means by (2) that $w(s,t) > 0$ implies $d(s,t) = 0$. Taking $x_{st} = w(s,t)$ for all $s,t \in S$, we see that $\{x_{st} : s,t \in S\}$ is a feasible solution to (MP). It follows that $\hat{d}(\mu, \eta) = 0$ since $\vee_{s,t\in S}(d(s,t) \wedge x_{st}) = 0$. This finishes the proof. ∎

*Proof of Theorem 3:* We first show the sufficiency. Assume that $d_R$ is a post-fixed point of $\Delta$. For any $(s,t) \in R$ and $a \in A$, if $s \xrightarrow{a} \mu$, then by Lemma 9 there exists some $\eta$ such that $t \xrightarrow{a} \eta$ and $\hat{d}_R(\mu, \eta) \leq d_R(s,t)$. In order to show that $R$ is a bisimulation, it remains to prove that $\mu(C) = \eta(C)$ for any equivalence class $C \in S/R$. In fact, we see that $\hat{d}_R(\mu, \eta) = 0$ since $d_R(s,t) = 0$ by the definition of $d_R$. Note that by Lemma 8 $d_R$ is a pseudo-ultrametric on $S$. It thus follows from Lemma 7 that $(\mu, \eta) \in \hat{R}$. As a result, we obtain by Lemma 6 that $\mu(C) = \eta(C)$ for any $C \in S/R$, as desired.

For the necessity, suppose that $R$ is a bisimulation. We now proves $d_R \preceq \Delta(d_R)$ by Lemma 9. Let $(s,t) \in S \times S$ and $a \in A$. If $(s,t) \notin R$, then $d_R(s,t) = 1$ by the definition of $d_R$ and the condition of Lemma 9 is naturally satisfied. On the other hand, if $(s,t) \in R$ and $s \xrightarrow{a} \mu$, then by the definition of bisimulation there exists some $\eta$ such that $t \xrightarrow{a} \eta$ and $\mu(C) = \eta(C)$ for all $C \in S/R$. The latter implies that $(\mu, \eta) \in \hat{R}$ by Lemma 6. Whence, we see by Lemma 7 that $\hat{d}_R(\mu, \eta) = 0$, which means that the condition of Lemma 9 holds. Therefore, $d_R$ is a post-fixed point of $\Delta$, finishing the proof of the theorem. ∎

*Proof of Theorem 4:* For the 'only if' part, suppose that $s \sim t$. Then there is a bisimulation $R$ that contains $(s,t)$. Thus, we get by Theorem 3 that $d_R$ is a post-fixed point of $\Delta$ with $d_R(s,t) = 0$. By the definition of $d_f$, we have that $d_R \preceq d_f$. Consequently, $d_f(s,t) \leq d_R(s,t) = 0$, which forces that $d_f(s,t) = 0$.

For the converse, consider the relation $R$ defined by $R = \{(s,t) \in S \times S : d_f(s,t) = 0\}$, namely, $(s,t) \in R$ if and only if $d_f(s,t) = 0$. Clearly, $R$ is an equivalence relation. It remains to show that this equivalence relation is a bisimulation. For any $(s,t) \in R$ and $a \in A$, if $s \xrightarrow{a} \mu$, we obtain by Lemma 9 that there exists some $\eta$ such that $t \xrightarrow{a} \eta$ and $\hat{d}_f(\mu, \eta) \leq d_f(s,t)$, because the greatest fixed point $d_f$ is also a post-fixed point of $\Delta$. As $(s,t) \in R$, we see that $d_f(s,t) = 0$ and thus $\hat{d}_f(\mu, \eta) = 0$, which implies that $(\mu, \eta) \in \hat{R}$ by Lemma 7. It follows from Lemma 6 that $\mu(C) = \eta(C)$ for all $C \in S/R$. This proves that $R$ is bisimulation, completing the proof. ∎

*Proof of Corollary 4:* As $s \sim s'$ and $t \sim t'$, we see by Theorem 4 that $d_f(s,s') = d_f(t,t') = 0$. Note that $d_f$ is a

$$\begin{aligned}
\hat{D}(\mu \wedge \mu', \eta \wedge \eta') &\leq \vee_{s|s',t|t' \in S \times S} \left[ D(s|s',t|t') \wedge z_{s|s',t|t'} \right] \\
&= \vee_{s,s',t,t' \in S} \left[ (d_f(s,t) \vee d_f(s',t')) \wedge (x_{st} \wedge y_{s't'}) \right] \\
&= \vee_{s,s',t,t' \in S} \left[ (d_f(s,t) \wedge x_{st} \wedge y_{s't'}) \vee (d_f(s',t') \wedge x_{st} \wedge y_{s't'}) \right] \\
&\leq \vee_{s,s',t,t' \in S} \left[ (d_f(s,t) \wedge x_{st}) \vee (d_f(s',t') \wedge y_{s't'}) \right] \\
&= \left[ \vee_{s,t \in S}(d_f(s,t) \wedge x_{st}) \right] \vee \left[ \vee_{s',t' \in S}(d_f(s',t') \wedge y_{s't'}) \right] \\
&= \hat{d}_f(\mu, \eta) \vee \hat{d}_f(\mu', \eta') \\
&\leq d_f(s,t) \vee d_f(s',t') \\
&= D(s|s',t|t').
\end{aligned}$$

---

pseudo-ultrametric. We thus get by (P3) that

$$\begin{aligned}
d_f(s,t) &\leq d_f(s,s') \vee d_f(s',t) \\
&= d_f(s',t) \\
&\leq d_f(s',t') \vee d_f(t',t) \\
&= d_f(s',t') \vee d_f(t,t') \\
&= d_f(s',t'),
\end{aligned}$$

namely, $d_f(s,t) \leq d_f(s',t')$. The converse can be proved in the same way. Thereby, $d_f(s,t) = d_f(s',t')$, as desired. ∎

*Proof of Theorem 5:* We only prove the first assertion; the second one can be proved similarly.

For (1), consider the parallel composition $(S \times S, A, \delta')$. Let us first define a function $D : (S \times S) \times (S \times S) \longrightarrow [0,1]$ by setting

$$D(s|s',t|t') = d_f(s,t) \vee d_f(s',t')$$

for any $s|s',t|t' \in S \times S$. It is easy to check that $D$ is a pseudo-ultrametric on $S \times S$. We claim that $D$ is a post-fixed point of $\Delta'$.

Let us verify the claim by using Lemma 9. For any given $(s|s',t|t') \in (S \times S) \times (S \times S)$ and $a \in A$, suppose that $s|s' \xrightarrow{a} \mu \wedge \mu'$. We need to show that there exists some $\eta \wedge \eta'$ such that $t|t' \xrightarrow{a} \eta \wedge \eta'$ and $\hat{D}(\mu \wedge \mu', \eta \wedge \eta') \leq D(s|s',t|t')$. We have to discuss four cases: $a \in A_s \backslash A_{s'}$, $a \in A_{s'} \backslash A_s$, $a \in A_s \cap A_{s'}$, and $a \notin A_s \cup A_{s'}$. We only go into the details of the third case; the other cases are simpler and can be proved in a similar way.

In the case of $a \in A_s \cap A_{s'}$, we get by the definition of parallel composition that there are $s \xrightarrow{a} \mu$ and $s' \xrightarrow{a} \mu'$. Because the greatest fixed point $d_f$ of $\Delta$ is a post-fixed point as well, by Lemma 9 there are $\eta$ and $\eta'$ such that $t \xrightarrow{a} \eta$ and $t' \xrightarrow{a} \eta'$. Moreover, we get that $\hat{d}_f(\mu, \eta) \leq d_f(s,t)$ and $\hat{d}_f(\mu', \eta') \leq d_f(s',t')$. It yields that $t|t' \xrightarrow{a} \eta \wedge \eta'$ and remains to verify that $\hat{D}(\mu \wedge \mu', \eta \wedge \eta') \leq D(s|s',t|t')$. Two subcases need to be considered: One is that both $\mu(S) = \eta(S)$ and $\mu'(S) = \eta'(S)$; the other is that either $\mu(S) \neq \eta(S)$ or $\mu'(S) \neq \eta'(S)$.

In the first subcase, we may assume that $\hat{d}_f(\mu, \eta)$ and $\hat{d}_f(\mu', \eta')$ are attained by $\{x_{st} : s,t \in S\}$ and $\{y_{s't'} : s',t' \in S\}$, respectively, that is, $\hat{d}_f(\mu,\eta) = \vee_{s,t \in S}(d_f(s,t) \wedge x_{st})$ and $\hat{d}_f(\mu',\eta') = \vee_{s',t' \in S}(d_f(s',t') \wedge y_{s't'})$. Then we have the following:

$$\begin{aligned}
\vee_{t \in S} x_{st} &= \mu(s), \quad \forall s \in S & (8) \\
\vee_{s \in S} x_{st} &= \eta(t), \quad \forall t \in S & (9) \\
\vee_{t' \in S} y_{s't'} &= \mu'(s'), \quad \forall s' \in S & (10) \\
\vee_{s' \in S} y_{s't'} &= \eta'(t'), \quad \forall t' \in S & (11)
\end{aligned}$$

For all $s|s',t|t' \in S \times S$, set $z_{s|s',t|t'} = x_{st} \wedge y_{s't'}$. For any $s|s' \in S \times S$, we find that

$$\begin{aligned}
\vee_{t|t' \in S \times S} z_{s|s',t|t'} &= \vee_{t,t' \in S}(x_{st} \wedge y_{s't'}) \\
&= \vee_{t \in S}[x_{st} \wedge (\vee_{t' \in S} y_{s't'})] \\
&\stackrel{(10)}{=} \vee_{t \in S}[x_{st} \wedge \mu'(s')] \\
&= (\vee_{t \in S} x_{st}) \wedge \mu'(s') \\
&\stackrel{(8)}{=} \mu(s) \wedge \mu'(s') \\
&= (\mu \wedge \mu')(s|s'),
\end{aligned}$$

namely, $\vee_{t|t' \in S \times S} z_{s|s',t|t'} = (\mu \wedge \mu')(s|s')$. On the other hand, for any $t|t' \in S \times S$, we have that

$$\begin{aligned}
\vee_{s|s' \in S \times S} z_{s|s',t|t'} &= \vee_{s,s' \in S}(x_{st} \wedge y_{s't'}) \\
&= \vee_{s \in S}[x_{st} \wedge (\vee_{s' \in S} y_{s't'})] \\
&\stackrel{(11)}{=} \vee_{s \in S}[x_{st} \wedge \eta'(t')] \\
&= (\vee_{s \in S} x_{st}) \wedge \eta'(t') \\
&\stackrel{(9)}{=} \eta(t) \wedge \eta'(t') \\
&= (\eta \wedge \eta')(t|t'),
\end{aligned}$$

i.e., $\vee_{s|s' \in S \times S} z_{s|s',t|t'} = (\eta \wedge \eta')(t|t')$. As a result, $\{z_{s|s',t|t'} : s|s',t|t' \in S \times S\}$ is a feasible solution to the mathematical programming problem (MP) defining $\hat{D}(\mu \wedge \mu', \eta \wedge \eta')$. Therefore, we get that $\hat{D}(\mu \wedge \mu', \eta \wedge \eta') \leq D(s|s',t|t')$ by the inequality at the top of this page.

In the second subcase, we see that $\hat{d}_f(\mu,\eta) = 1$ or $\hat{d}_f(\mu',\eta') = 1$. It follows that $d_f(s,t) = 1$ or $d_f(s',t') = 1$, since $\hat{d}_f(\mu,\eta) \leq d_f(s,t)$ and $\hat{d}_f(\mu',\eta') \leq d_f(s',t')$. Therefore, we get that $D(s|s',t|t') = d_f(s,t) \vee d_f(s',t') = 1 \geq \hat{D}(\mu \wedge \mu', \eta \wedge \eta')$. This completes the proof of the claim.

Based on the claim, it follows by the definition of $d'_f$ that $D \preceq d'_f$, which means that

$$\begin{aligned}
d'_f(s_1|s_2,t_1|t_2) &\leq D(s_1|s_2,t_1|t_2) \\
&= d_f(s_1,t_1) \vee d_f(s_2,t_2) \\
&= \epsilon_1 \vee \epsilon_2.
\end{aligned}$$

Consequently, the first assertion holds, finishing the proof. ∎

## REFERENCES

[1] D. Park, "Concurrency and automata on infinite sequences," in *Proceedings of the 5th GI-Conference on Theoretical Computer Science*, Lect. Notes Comput. Sci., G. Goos and J. Hartmanis, Eds. Springer Berlin, 1981, vol. 104, pp. 167–183.

[2] R. Milner, *Communication and Concurrency*. Englewood Cliffs, New Jersey: Prentice-Hall, 1989.

[3] D. Sangiorgi, "On the origins of bisimulation and coinduction," *ACM Trans. Program. Lang. Syst.*, vol. 31, no. 4, pp. 1–41, 2009.

[4] T. Petković, "Congruences and homomorphisms of fuzzy automata," *Fuzzy Sets Syst.*, vol. 157, pp. 444–458, 2006.

[5] P. Buchholz, "Bisimulation relations for weighted automata," *Theor. Comput. Sci.*, vol. 393, pp. 109–123, 2008.

[6] D. D. Sun, Y. M. Li, and W. W. Yang, "Bisimulation relations for fuzzy finite automata," *Fuzzy Syst. Math.*, vol. 23, pp. 92–100, 2009, (in Chinese).

[7] Y. Cao, G. Chen, and E. Kerre, "Bisimulations for fuzzy-transition systems," *IEEE Trans. Fuzzy Syst.*, vol. 19, no. 3, pp. 540–552, Jun. 2011.

[8] Y. Cao and Y. Ezawa, "Modeling and specification of nondeterministic fuzzy discrete-event systems," Dec. 2010, submitted to *IEEE Trans. Automat. Contr.*, revised.

[9] K. G. Larsen and A. Skou, "Bisimulation through probabilistic testing," *Inform. Comput.*, vol. 94, pp. 1–28, 1991.

[10] M. Ćirić, A. Stamenković, J. Ignjatović, and T. Petković, "Factorization of fuzzy automata," in *Fundamentals of Computation Theory*, Lect. Notes Comput. Sci., E. Csuhaj-Varju and Z. Ésik, Eds., vol. 4639. Heidelberg: Springer, 2007, pp. 213–225.

[11] J. Ignjatović, M. Ćirić, and S. Bogdanović, "Fuzzy homomorphisms of algebras," *Fuzzy Sets Syst.*, vol. 160, no. 16, pp. 2345–2365, 2009.

[12] M. Ćirić, A. Stamenković, J. Ignjatović, and T. Petković, "Fuzzy relation equations and reduction of fuzzy automata," *J. Comput. Syst. Sci.*, vol. 76, no. 7, pp. 609–633, Nov. 2010.

[13] M. Ćirić, J. Ignjatović, N. Damljanović, and M. Bašić, "Bisimulations for fuzzy automata," *Fuzzy Sets Syst.*, 2011, in press, corrected proof, doi: 10.1016/j.fss.2011.07.003.

[14] A. Stamenković, M. Ćirić, and J. Ignjatović, "Reduction of fuzzy automata by means of fuzzy quasi-orders," *Arxiv preprint arXiv:1102.5451*, 2011.

[15] M. Ćirić, J. Ignjatović, I. Jančić, and N. Damljanović, "Algorithms for computing the greatest simulations and bisimulations between fuzzy automata," *Arxiv preprint arXiv:1103.5078*, 2011.

[16] J. Desharnais, R. Jagadeesan, V. Gupta, and P. Panangaden, "The metric analogue of weak bisimulation for probabilistic processes," in *Proceedings of the 17th Annual IEEE Symposium on Logic in Computer Science (LICS'02)*. IEEE Press, 2002, pp. 413–422.

[17] J. Desharnais, V. Gupta, R. Jagadeesan, and P. Panangaden, "Metrics for labelled markov processes," *Theor. Comput. Sci.*, vol. 318, no. 3, pp. 323–354, 2004.

[18] N. Ferns, P. Panangaden, and D. Precup, "Metrics for finite markov decision processes," in *Proceedings of the Twentieth Conference Annual Conference on Uncertainty in Artificial Intelligence (UAI-04)*. Arlington, Virginia: AUAI Press, 2004, pp. 162–169.

[19] ——, "Metrics for markov decision processes with infinite state spaces," in *Proceedings of the Twenty-First Conference Annual Conference on Uncertainty in Artificial Intelligence (UAI-05)*. Arlington, Virginia: AUAI Press, 2005, pp. 201–208.

[20] F. van Breugel and J. Worrell, "A behavioural pseudometric for probabilistic transition systems," *Theor. Comput. Sci.*, vol. 331, no. 1, pp. 115–142, 2005.

[21] Y. Deng, T. Chothia, C. Palamidessi, and J. Pang, "Metrics for action-labelled quantitative transition systems," *Electr. Notes Theor. Comput. Sci.*, vol. 153, no. 2, pp. 79–96, 2006.

[22] D. Repovš, A. Savchenko, and M. Zarichnyi, "Fuzzy prokhorov metric on the set of probability measures," *Fuzzy Sets Syst.*, vol. 175, no. 1, pp. 96–104, 2011.

[23] C. A. R. Hoare, *Communicating Sequential Processes*. Englewood Cliffs, NJ: Prentice-Hall, 1985.

[24] L. A. Zadeh, "Fuzzy sets," *Inform. Contr.*, vol. 8, pp. 338–353, 1965.

[25] Y. Cao and Y. Ezawa, "Nondeterministic fuzzy automata," Nov. 2010, arxiv preprint arXiv:1012.2162, submitted to *Inform. Sci.*, revised.

[26] B. Leclerc, "Description combinatoire des ultramétriques," *Math. Sci. Humaines*, vol. 73, pp. 5–37, 127, 1981.

[27] A. J. Lemin, "Spectral decomposition of ultrametric spaces and topos theory," *Topology Proc.*, vol. 26, no. 2, pp. 721–739, 2001–2002.

[28] N. Uglešić, "On ultrametrics and equivalence relations–duality," *Intern. Math. Forum*, vol. 5, no. 21, pp. 1037–1048, 2010.

[29] A. M. Vershik, "Kantorovich metric: initial history and little-known applications," *J. Math. Sci.*, vol. 133, no. 4, pp. 1410–1417, 2006.

[30] Y. Deng and W. Du, "The kantorovich metric in computer science: a brief survey," *Electr. Notes Theor. Comput. Sci.*, vol. 253, no. 3, pp. 73–82, 2009.

[31] A. Tarski, "A lattice-theoretical fixpoint theorem and its applications," *Pac. J. Math.*, vol. 5, no. 2, pp. 285–309, 1955.

[32] L. A. Zadeh, "Similarity relations and fuzzy orderings," *Inform. Sci.*, vol. 3, no. 2, pp. 177–200, 1971.

[33] B. Jonsson and K. G. Larsen, "Specification and refinement of probabilistic processes," in *Proceedings of the 6th Annual IEEE Symposium on Logic in Computer Science (LICS'91)*. IEEE Press, 1991, pp. 266–277.

[34] J. Desharnais, V. Gupta, R. Jagadeesan, and P. Panangaden, "Metrics for labeled markov systems," in *CONCUR'99*, Lect. Notes Comput. Sci., J. C. M. Baeten and S. Mauw, Eds., vol. 1664. London, UK: Springer, 1999, pp. 258–273.

[35] F. van Breugel, C. Hermida, M. Makkai, and J. Worrell, "Recursively defined metric spaces without contraction," *Theor. Comput. Sci.*, vol. 380, no. 1–2, pp. 143–163, 2007.

[36] M. Ying, "Bisimulation indexes and their applications," *Theor. Comput. Sci.*, vol. 275, no. 1-2, pp. 1–68, 2002.

[37] A. Aldini, M. Bravetti, and R. Gorrieri, "A process-algebraic approach for the analysis of probabilistic noninterference," *J. Comput. Secur.*, vol. 12, no. 2, pp. 191–245, 2004.

[38] S. Tini, "Non-expansive ϵ-bisimulations for probabilistic processes," *Theor. Comput. Sci.*, vol. 411, no. 22-24, pp. 2202–2222, 2010.

[39] Y. Cao, "Reliability of mobile processes with noisy channels," *IEEE Trans. Comput.*, 2011, in press, doi: 10.1109/TC.2011.147.

[40] D. Dubois and H. Prade, *Fuzzy Sets and Systems: Theory and Applications*. New York: Academic Press, 1980.

[41] M. Ying, "A formal model of computing with words," *IEEE Trans. Fuzzy Syst.*, vol. 10, no. 5, pp. 640–652, Oct. 2002.

[42] Y. Cao, M. Ying, and G. Chen, "Retraction and generalized extension of computing with words," *IEEE Trans. Fuzzy Syst.*, vol. 15, no. 6, pp. 1238–1250, Dec. 2007.

[43] Y. Cao and G. Chen, "A fuzzy petri-nets model for computing with words," *IEEE Trans. Fuzzy Syst.*, vol. 18, pp. 486–499, Jun. 2010.

[44] B. A. Davey and H. A. Priestley, *Introduction to Lattices and Order*. Cambridge: Cambridge University Press, 1990.